\documentclass[runningheads]{llncs}

\newif\ifcameraready
\newif\ifbody
\newif\ifappendix
\newif\iflivetikz

\camerareadytrue
\bodytrue
\appendixtrue

\ifcameraready
\fi

\usepackage{graphicx}
\usepackage{comment}
\usepackage{amsmath,amssymb} 
\usepackage{color}
\usepackage[figuresleft]{rotating}
\usepackage{subcaption}

\unless\ifcameraready
\usepackage{ruler}
\usepackage[width=122mm,left=12mm,paperwidth=146mm,height=193mm,top=12mm,paperheight=217mm]{geometry}
\fi

\iflivetikz
\usepackage{tikz}
\usetikzlibrary{calc,positioning}
\usetikzlibrary{shapes,arrows}
\usetikzlibrary{backgrounds}
\usepackage{pgfplots}
\fi

\newcommand{\fulltitle}{Blind Image Restoration without Prior Knowledge}

\newcommand{\eg}{{\it e.g.}, }
\newcommand{\ie}{{\it i.e.}, }

\newcommand{\sidechain}{{SNSC}}
\newcommand{\cbdnethalf}{{CBDNet$_{1/2}$}}

\newcommand{\submod}[1]{\emph{#1}}
\newcommand{\relevance}{\submod{relevance}}
\newcommand{\validity}{\submod{validity}}
\newcommand{\estimate}{\submod{estimate}}

\iflivetikz
\input{SelfNormedCNN_900_TikzStyle}
\fi

\begin{document}
\pagestyle{headings}
\mainmatter
\def\ECCVSubNumber{425}  

\ifbody
\title{\fulltitle} 
\else
\title{\fulltitle\ -- Appendices}
\fi

\unless\ifcameraready
\titlerunning{ECCV-20 submission ID \ECCVSubNumber} 
\authorrunning{ECCV-20 submission ID \ECCVSubNumber} 
\author{Anonymous ECCV submission}
\institute{Paper ID \ECCVSubNumber}
\fi

\ifcameraready
\titlerunning{\fulltitle}
%
\author{Noam Elron \and Shahar S. Yuval \and Dmitry Rudoy \and Noam Levy}
\authorrunning{N. Elron et al.}
%
\institute{Intel Corporation\\
\email{noam.elron@intel.com}}
\fi
\maketitle

\ifbody
\begin{abstract}
Many image restoration techniques are highly dependent on the degradation used during training, and their performance declines significantly when applied to slightly different input. 
Blind and universal techniques attempt to mitigate this by producing a trained model that can adapt to varying conditions. 
However, blind techniques to date require prior knowledge of the degradation process, and assumptions regarding its parameter-space. 
In this paper we present the \emph{Self-Normalization Side-Chain} (\sidechain), a novel approach to blind universal restoration in which no prior knowledge of the degradation is needed. 
This module can be added to any existing CNN topology, and is trained along with the rest of the network in an end-to-end manner.
The imaging parameters relevant to the task, as well as their dynamics, are deduced from the variety in the training data.
We apply our solution to several image restoration tasks, and demonstrate that the \sidechain\ encodes the degradation-parameters, improving restoration performance.
	

\end{abstract}

\begin{figure}[!b]
	\centering
	\iflivetikz
		\input{SelfNormedCNN_909a_RESIDEResults}
	\else
		\includegraphics[trim=5 0 0 0,clip]{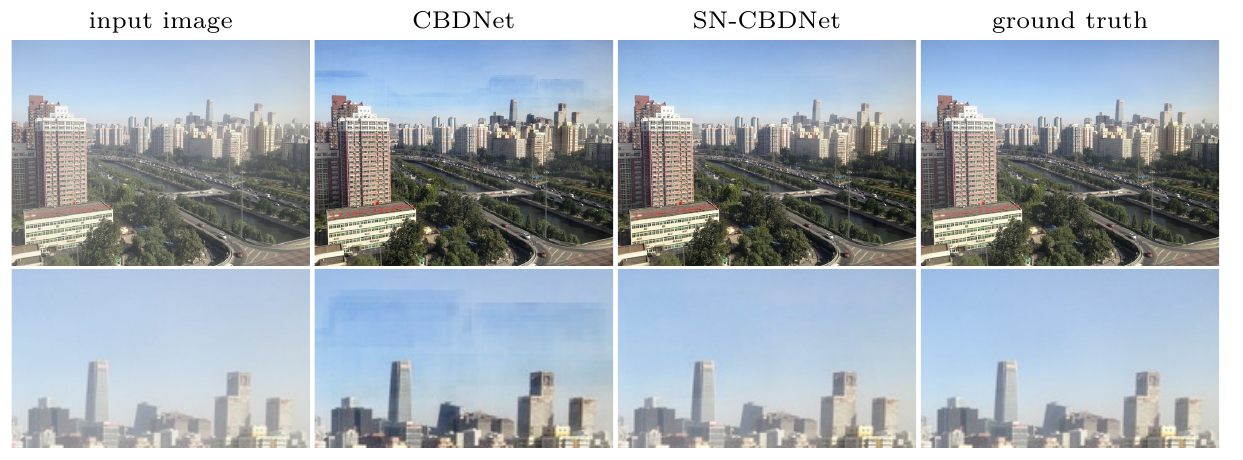}
	\fi
	\caption{Image dehazing result (Sec.~\ref{sec:dehaze}).
		CBDNet~\cite{CBDNet} serves as a baseline CNN, and \mbox{SN-CBDNet} is a variant incorporating our novel component.
		CBDNet generates large-scale artefacts (\eg the sky), whereas our result does not.}
	\label{fig:dehaze_results_cover}
\end{figure}

\section{Introduction}
\label{sec:intro}

Image restoration is a family of inverse problems for obtaining a high quality image from a corrupted input image.
Corruption may occur due to the image-capture process (\eg noise, lens blur), post-processing (\eg {JPEG} compression), or photography in non-ideal conditions (\eg haze, motion blur).

The state of the art in image restoration techniques is dominated by discriminative deep-learning models.
The first generation of these techniques~\cite{DnCNN,SRCNN,DeepDeconv} provided high-quality results but lacked flexibility -- the models were trained to remedy a specific corruption (\eg remove additive noise with prescribed variance), and their performance deteriorated dramatically when applied to images which had undergone a slightly different degradation process.
This well-documented sensitivity to the degradation-parameters~\cite{FFDNet} makes their application impractical.

To remedy this shortcoming, much research is aimed at blind and universal deep image restoration~\cite{FFDNet,GaussianNoiseLevelLearning,DynamicResidualDenseNetworks,BlindDeblurSurvey,BlindSuperResIterative,SRMD}.
An algorithm is \emph{universal} when a single model can handle a wide range of degradation-parameters (\eg a range of noise-levels), and \emph{blind} when it requires no external cues of the degradation-parameters for inference.
Existing approaches to blind processing include modifications to the model architecture~\cite{FFDNet,GaussianNoiseLevelLearning,DynamicResidualDenseNetworks,CBDNet}, to the loss-function~\cite{DeepImagePriors}, and to the training procedure~\cite{OnDemand}.
Some techniques use iterative inference, wherein the quality of output is evaluated and improved through iteration~\cite{plugandplay,BlindSuperResIterative}.

However, all blind techniques to date necessitate prior knowledge of the degradation process, and assumptions regarding its parameter-space.
In particular, some methods require quantization of the degradation-parameters~\cite{OnDemand,DeepImagePriors} or iterative corrections to the parameter-estimate~\cite{BlindSuperResIterative}, making them infeasible when the degradation-space is complex (multi-dimensional).
Having to employ prior knowledge undermines the elegant promise of supervised deep-learning -- a dataset of labelled exemplars is no longer sufficient to train the model without further guidance.

Mobile platforms nowadays include deep inference engines in hardware.
Because cost effective design of such accelerators is crucial, many employ a fixed-topology (for power efficiency) with programmable weights (enabling flexible usage on different tasks).
Since the topology cannot be adapted to the requirements of each specific task, enabling blind processing with no prior assumptions regarding the use-case is invaluable for such design.

In this work we present a novel approach to blind universal restoration, in which no prior knowledge of the degradation is needed.
Our method consists of a small trainable module that can be attached to any convolutional architecture (Figure~\ref{fig:side_chain_net}) -- a \emph{Self-Normalization Side-Chain} (\sidechain).
The main branch of the CNN and the side-chain are trained jointly in an end-to-end manner, with no extra supervision needed.
We demonstrate that an encoding of the degradation-parameters is deduced from the diversity in the training data itself.
At inference, the \sidechain\ can (a) estimate the current values of the degradation-parameters from the input image, and (b) tune the processing in the main branch to the current degradation.

\begin{figure}
	\centering
	\iflivetikz
		\input{SelfNormedCNN_901_SideChain}
	\else
		\includegraphics[trim=0 0 0 0,clip]{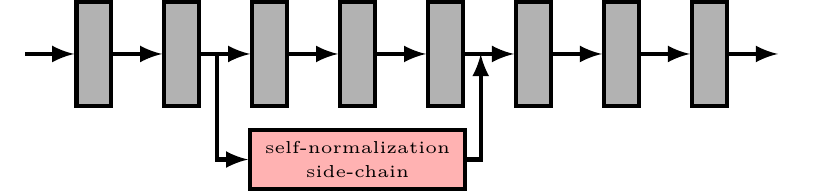}
	\fi
	\caption{The Self-Normalization Side-Chain attached to a simple feed-forward CNN.}
	\label{fig:side_chain_net}
\end{figure}

The \sidechain\ performs two operations.
First a \emph{Global Weighted-Average Pooling} (GWAP) layer aggregates frame-wide measurements into a compact representation.
This global information is then \emph{modulated} by a measure of its relevance to the specific location in the image, and fed back into the main branch of the CNN.
The measurements, the averaging-weights, and the relevance-metric are all learned through end-to-end training.
A detailed description is provided in Section~\ref{sec:method}.

The main contributions of this work are:
\begin{itemize}
	\item A generic topological component, designed for estimating and using frame-wide parameters.
	This component is trained end-to-end, requiring no prior knowledge of the task and the degradation-space.
	\item Introducing global \emph{weighted}-average pooling, and applying it for the above purpose.
	\item Multiplicative modulation of global information in order to adapt to local conditions.
	\item A very-low-cost mechanism for modifying CNN output.
\end{itemize}

\section{Related Work}
\label{sec:related}

\subsubsection{Blind Deep Image Restoration}

Much research has been dedicated to creating blind and universal image restoration methods.
These techniques can be roughly classified into several categories:
\begin{itemize}
\item Universal but non-blind -- degradation-parameters are estimated externally and used as an auxiliary input to the model~\cite{FFDNet,cai2016dehazenet}.
\item Deep control, wherein the architecture of the model includes a deep component which either estimates the degradation-parameters~\cite{GaussianNoiseLevelLearning,BlindSuperResIterative,CBDNet,Li2019lapnet,li2017aodnet,zhang2018dcpdn} or adapts the topology of the model~\cite{DynamicResidualDenseNetworks}.
\item Elaborate training procedures~\cite{OnDemand} or loss-functions~\cite{DeepImagePriors} for learning an implicit estimate of the degradation-parameters.
\item Iterative procedures for obtaining an improved output at every iteration~\cite{plugandplay,BlindSuperResIterative}.
\end{itemize}
All of the methods above necessitate \emph{prior knowledge} of the degradation and its parameter-space.
Many of these methods are tailored for a specific image restoration task~\cite{cai2016dehazenet,BlindSuperResIterative}, and cannot be easily generalized to other tasks.
In addition, some approaches require quantization of the degradation-parameters~\cite{OnDemand,DeepImagePriors}, making them infeasible when the degradation-space is complex (multi-dimensional).

\subsubsection{Control in Deep Image Restoration}

Another challenge in making deep inference methods for real-world image restoration applications is parametric processing -- users want to adjust the output image to their tastes, whereas a traditional CNN can only supply a deterministic result.
Several papers have been dedicated to parametrization of deep inference~\cite{DeepInterp,MetaSR,He_2019_CVPR,Shoshan_2019_ICCV}, wherein a large percentage of the weights or layers are manipulated in order to achieve variations in the output image.
In~\cite{ParametrizationEval} the authors review a variety of parametrization approaches.
In Section~\ref{sec:flexible} we show that the \sidechain\ has the potential to provide parametrization capabilities with manipulation of only a tiny fraction of the CNN weights.

\subsubsection{Global Average Pooling}

The Global Average Pooling (GAP) layer was pioneered as an alternative to fully-connected layers in the design of image classification networks~\cite{Lin2013NetworkIN}.
Since then it has been popular in channel-attention mechanisms~\cite{CBAM,SqueezeExcitation1,SqueezeExcitation2}, where it is used quantify the relative importance of each channel.

In the context of image restoration tasks, GAP layers are used in~\cite{gharbi2017deep,Colorization,DeepImagePriors} to produce global features of the image.
In~\cite{gharbi2017deep} and~\cite{DeepImagePriors} the GAP layers are placed in the later stages of the processing pipe-line, leaving little capacity for utilizing the global features in the processing.
In~\cite{Colorization} global averaging is effectively extreme downscaling as a part of a complex multi-scale architecture.
All of the above contributions use simple averaging.


\section{Self-Normalization Side-Chain}
\label{sec:method}

The main contribution of this work is the design of the Self-Normalization Side-Chain.
This module, which can be added to any existing architecture, provides intermediate layers in the CNN with information which is aggregated over a much larger receptive field than would otherwise be available to them, thereby allowing them to adapt to global imaging conditions.
The \sidechain\ contains four trainable components (Figure~\ref{fig:side_chain_internal}) -- channel-compression to $S$ internal channels followed by three shallow CNNs, typically 1-3 layers deep, \estimate, \validity, and \relevance.

\begin{figure}
	\centering
	\iflivetikz
	\input{SelfNormedCNN_902_SideChainInternal}
	\else
	\includegraphics[trim=0 0 0 0,clip,width=0.92\textwidth]{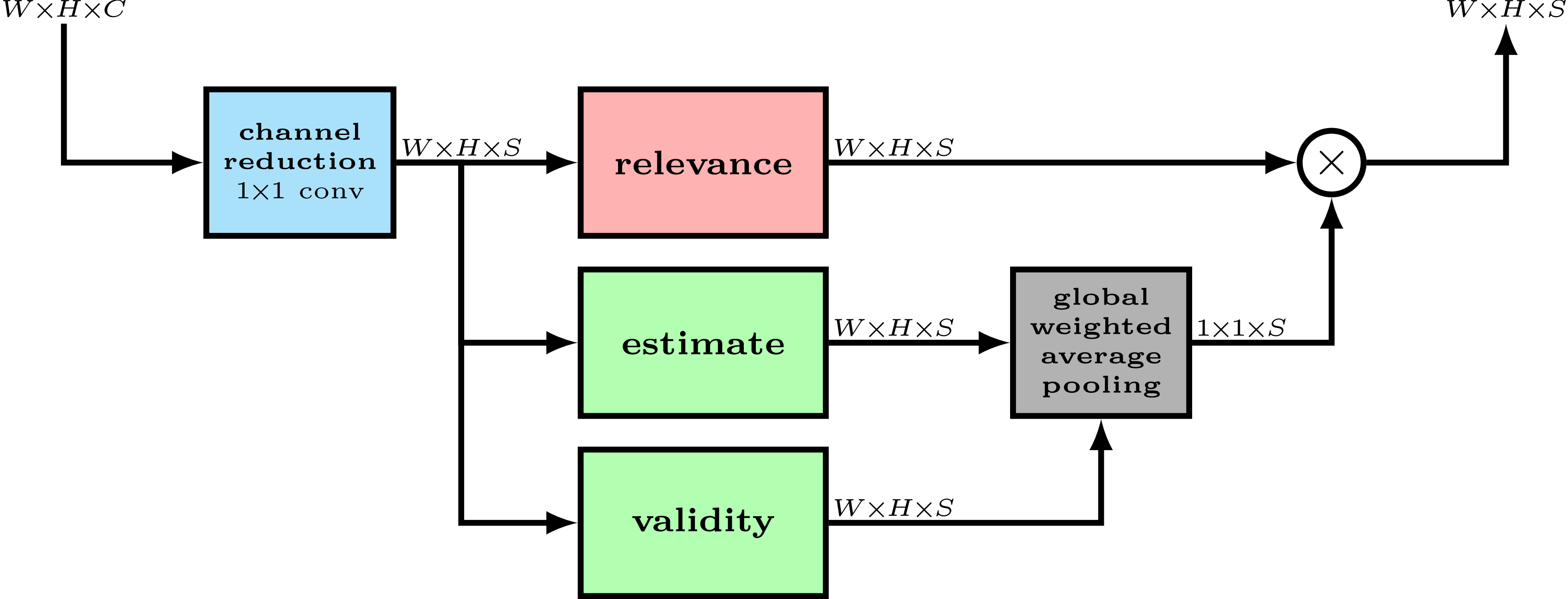}
	\fi
	\caption{Internal structure of the Self-Normalization Side-Chain.
		The input undergoes channel-compression (trainable point-wise convolution) and then fed into three shallow CNNs (typically 1-3 layers deep) -- \estimate\ and \validity\ produce the values and weights fed into the GWAP layer.
		The weighted-averages are modulated by the local \relevance\ before being returned into the main processing branch.}
	\label{fig:side_chain_internal}
\end{figure}

The computation is comprised of two steps.
The first is Global Weighted Average Pooling, where both the pooled features and the averaging-weights are learned in training.
The \estimate\ component produces local measurements related to the degradation.
The \validity\ component assigns a score to each measurement that serves as the weight in the frame-wide averaging.
We delve deeper into the mechanics of the GWAP operation in Section~\ref{sec:gwap}.
The output of the pooling operation is a \emph{state-vector}, which is a compact characterization of the entire frame.

The second element of the design is multiplicative modulation of the global state by a local value before injecting it back into the main CNN.
This permits the SNCN to modify the output based on the local \relevance\ of each state-element.
This is also a trained component.
Adapting the application of the degradation-parameters to local conditions is crucial when the degradation is signal-dependent~\cite{Foi1}.
A beneficial by-product of relevance modulation is that different processing effects can be obtained (Section~\ref{sec:flexible}).

\subsection{Degradation Parameters in Classic Restoration Algorithms}
\label{sec:intuition}

To illustrate the role of degradation-parameters in image restoration, let us examine the bilateral filter~\cite{bilateral1,bilateral2}.
The ``bilateral principle'' is ubiquitous and has variants for many applications, and is also an archetype for many classic data-dependent processes.
We will consider the original noise-reduction formulation wherein for a given underlying signal $x$ contaminated by zero-mean noise $y=x+n$, the restoration operator is
\begin{equation}
\hat{x}(t) = G(t) \cdot \int y(\tau) \cdot W_s \left( | t - \tau | \right) \cdot W_i \left( \| y(t) - y(\tau) \| \right) d\tau
\end{equation}
$W_s$ is a spatial-proximity profile, and $W_i$ is an intensity-proximity profile.
Both are multiplied and used as a weight for averaging input samples to produce the output.
$G(t)$ is the inverse of the sum of the weights required for normalization.

The profiles $W_s$ and $W_i$ are non-negative and tend to be monotonically decreasing over the positive interval $\tau \geq 0$, where the common choice is Gaussians of widths $\sigma_s$ and $\sigma_i$ respectively.
The intensity-proximity term in the averaging-weight ensures that the majority of contributions to each output sample come from ``similar'' samples in the input signal -- the averaging does not mix different signal-features and therefore salient discontinuities (edges, textures, etc.) are preserved.

Conceptually, the profile $W_i$ can be split into three regimes (see Fig.~\ref{fig:weightprofiles}) -- low values of the intensity difference $\|y(t)-y(\tau)\|$ are considered similar and receive a large averaging-weight, high values are divergent and get a low weight, and a transition region assures continuous behaviour.
But determining at what values of $\|y(t)-y(\tau)\|$ to transition from similar to divergent depends on the level of noise in the signal $y$ (See Fig.~\ref{fig:noisysignal}).
Transitioning too late gives large weights to pixels on both sides of the edge, smearing the salient feature;
transitioning too early reduces the denoising power of the operation overall, resulting in residual noise.
Thus, for effective bilateral denoising the regime transition in $W_i$ must be proportional to the noise-level $\sigma_i \propto \sigma_n$.
In other words, in order to perform adequately, the algorithm must be supplied an estimate of the degradation-parameter $\sigma_n$ (or produce one internally), and \emph{normalize} its measurements based on this value.

\begin{figure}[ht]
	\begin{subfigure}[t]{.49\columnwidth}
		\centering
		\iflivetikz
		\input{SelfNormedCNN_906_WeightProfiles}
		\else
		\includegraphics[trim=0 0 0 0,clip,width=0.9\columnwidth]{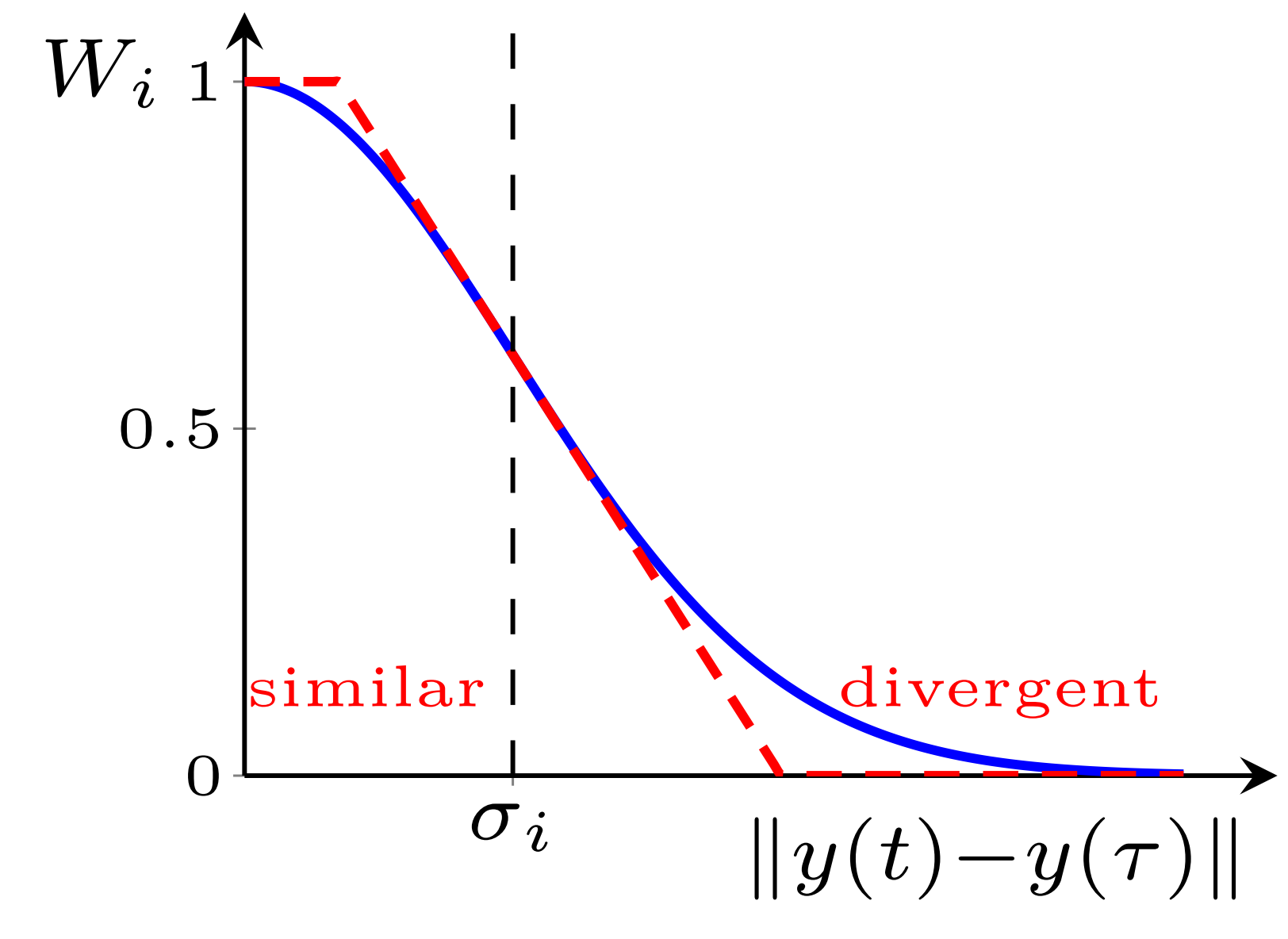}
		\fi
		\caption{Gaussian intensity proximity profile $W_i$ (solid blue) with ``three regime approximation'' (dashed red)}
		\label{fig:weightprofiles}
	\end{subfigure}
	\begin{subfigure}[t]{.49\columnwidth}
		\centering
		\iflivetikz
			\input{SelfNormedCNN_907_StepWithNoise}
		\else
			\includegraphics[trim=0 0 0 0,clip,width=0.9\columnwidth]{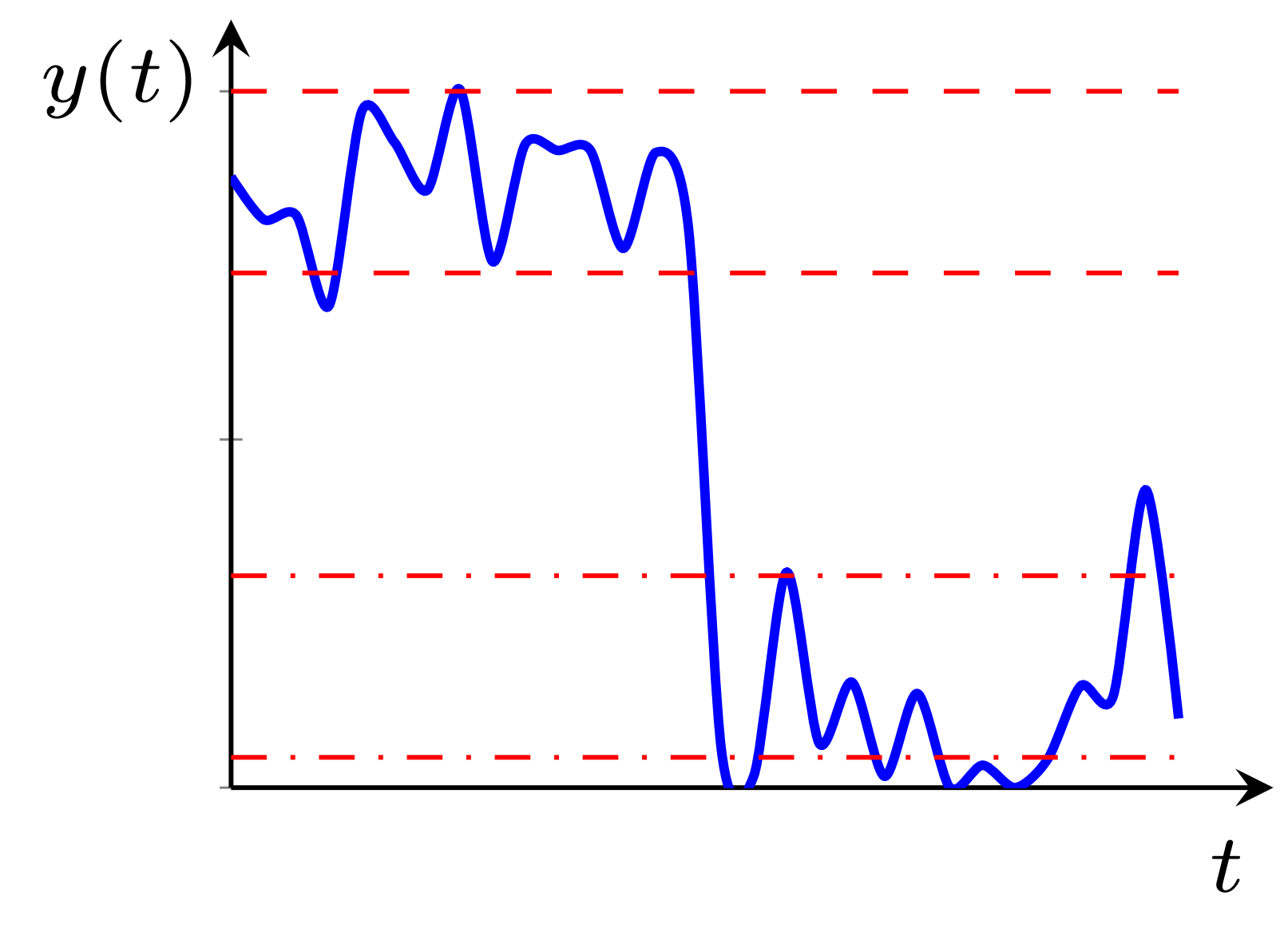}
		\fi
		\caption{Step function with additive noise}
		\label{fig:noisysignal}
	\end{subfigure}
	\caption{Illustrations of bilateral filter dynamics.}
	\label{fig:intuition}
\end{figure}

The bilateral filter is just an example -- dependence on imaging conditions is a property of the restoration task itself, and not so much the requirement of a specific algorithm.
We argue that this is also valid for data-driven methods, such as CNNs -- for high-quality restoration, the global degradation-parameters must be either supplied or estimated internally.

\subsection{Global Weighted Average Pooling}
\label{sec:gwap}

The challenge in estimating imaging parameters is that many image regions are unsuitable, and cannot be used in the estimation.
For example, noise level is estimated by collecting statistics of the local variance across a predefined area.
If the entire measurement area is highly textured, the variance is dominated by the texture, and the noise-level cannot be evaluated.
On the other hand, blur-width is best estimated by measuring the widths of salient edges, and cannot be reliably estimated when none are present.

Thus, robust estimation requires a very large receptive field, as it is the only way to ensure that a region suitable for estimation is ``in sight''.
As a consequence, CNNs can only gain access to such estimates late in the data-flow (after enough spatial information has been aggregated), and the capacity of the remaining layers is not sufficient to make effective use of them.
In contrast, our approach integrates frame-wide statistics early in the flow by attaching the \sidechain\ close to the input of the CNN, thereby obtaining a large \emph{effective} receptive field.

We ensure that only the suitable parts of the image are used in the parameter estimation by relying on \emph{weighted} averaging rather than simple averaging.
The weights are produced by the \validity\ component
\begin{equation}
\text{state}(s) = \frac{\sum_{x,y} \text{estimate}(x,y,s)\cdot\text{validity}(x,y,s)}{\sum_{x,y} \text{validity}(x,y,s) + \epsilon}.
\end{equation}
The output of the \validity\ component is activated using a ReLU, thus many of the weights are zeros.
In effect, it performs extremely conservative outlier rejection.
In Section~\ref{sec:complex_analysis} we demonstrate that the \sidechain\ makes use of only a small fraction of the potential measurements -- all but the most indicative measurements are discarded, making the estimation extremely robust.


\section{Experiments}
\label{sec:experiments}

We demonstrate the performance of the \sidechain\ in two tasks.
The first is restoration of images degraded by complex artificial blur and noise (Section~\ref{sec:complex}).
We use this use-case to gain insight into the inner workings of the \sidechain.
In Section~\ref{sec:dehaze} we test performance on image dehazing -- a very different restoration task, but one that relies heavily on robust estimation of global imaging conditions.
We show that in both tasks the \sidechain\ contributes to improved qualitative and quantitative results.
Finally, in Section~\ref{sec:flexible}, we demonstrate the potential of using the \sidechain\ to create a control mechanism for CNNs.

Our CNN architecture of choice is CBDNet~\cite{CBDNet}, which is a state-of-the-art image-denoising CNN.
Originally, CBDNet consists of two sub-modules, the first produces an estimate of the noise-characteristics from the image, and the second uses this estimate to perform image denoising.
In~\cite{CBDNet}, the estimation sub-module is trained using ground-truth maps of the noise variance, hence prior knowledge of both the task and the degradation-space is required.

In this work we use only the denoising sub-module of CBDNet, which is a U-Net design.
It has 6.76M trained parameters, and a receptive field of 96 pixels.
We also create SN-CBDNet, which is the same U-Net with a self-normalization side-chain running in parallel to the third and fourth layers of the input (full-resolution) encoder
\ifappendix
(see Figure~\ref{fig:sn_cbdnet} in Appendix~\ref{sec:sn_cbdnet}).
\else
(a detailed diagram is given in the appendices).
\fi
We demonstrate that the SNCN produces an implicit estimate of the degradation-parameters, learned solely from the training data.
The estimate propagates downstream, enabling a higher quality of image restoration.
The side-chain has an $S=8$ dimensional state vector, and each of the three internal components has just two convolutional layers (each consisting of $3\times3$ convolution, batch-normalization and ReLU activation), giving a total of 8.7K trained parameters --  a negligible overhead.

We also introduce a lighter variant \cbdnethalf\, which has identical topology as CBDNet, but with half the number of channels in all layers, \ie it has an identical receptive field but only one quarter of the trained parameters (1.69M).
We choose channel reduction as a slimming technique, while keeping the topology identical, so that both CNNs have access to similar spatial information at all stages of the processing stream.
The side-chain of the self-normalized counterpart SN-\cbdnethalf, is identical -- the state dimension is also $S=8$.

To summarize, we use four models in the experiments below: the original CBDNet from~\cite{CBDNet} and its self-normalized counterpart SN-CBDNet, and the reduced-capacity \cbdnethalf\ and SN-\cbdnethalf.

\subsection{Complex Artificial Degradation}
\label{sec:complex}

We now introduce a complex image-degradation scenario, and demonstrate that the \sidechain\ can estimate multiple unknown degradation-parameters without prior knowledge.
This estimate of the degradation-parameters enables improved image restoration.

A high-quality image is assumed to have undergone blurring by an anisotropic Gaussian kernel, followed by additive noise with signal-dependent variance~\cite{Foi1}.
Formally, the degraded image $I_{\text{corrupt}}$ is given by
\begin{equation} \label{eq:blurnoise}
\begin{split}
I_{\text{blurred}} &= b * I_{\text{groundtruth}} \\
I_{\text{corrupt}} &= I_{\text{blurred}} + N_{\sigma(I_{\text{blurred}})}
\end{split}
\end{equation}
where $b$ is a normalized Gaussian kernel $b(x,y) = \frac{1}{2\pi \sqrt{\left| \Lambda \right|}} \exp \{ -\frac{1}{2} \left( \begin{smallmatrix} x \\ y \end{smallmatrix} \right)^T \Lambda^{-1} \left( \begin{smallmatrix} x \\ y \end{smallmatrix} \right) \}$ with covariance
\begin{equation}
\Lambda=\begin{pmatrix} w_x^2 & \rho w_x w_y \\ \rho w_x w_y & w_y^2 \end{pmatrix}.
\end{equation}
The additive noise $N_{\sigma(I_{\text{blurred}})}$ is zero-mean Gaussian noise with local variance
\begin{equation}
\sigma (x,y) = \left( 1 - L_{\text{blurred}}(x,y) \right) \cdot \sigma_{\text{dark}} + L_{\text{blurred}}(x,y) \cdot \sigma_{\text{bright}},
\end{equation}
where $L_{\text{blurred}}$ is the local luminosity of $I_{\text{blurred}}$.
All in all, the degradation-space has five parameters $w_x$, $w_y$, $\rho$, $\sigma_{\text{dark}}$, and $\sigma_{\text{bright}}$.

We create a dataset of natural images corrupted using the model in equation~\eqref{eq:blurnoise}.
Each image is degraded using random parameters (we assume images are normalized to the range $[0,1]$)
\begin{equation} \label{eq:blurnoiseparams}
\begin{split}
w_x, w_y&\sim U[0,2] \\
\rho&\sim U[-1, 1] \\
\sigma_{\text{dark}},\sigma_{\text{bright}}&\sim U[0,0.15]
\end{split}
\end{equation}

We train all four CNNs described in Section~\ref{sec:experiments} to restore images degraded according to~\eqref{eq:blurnoise} and~\eqref{eq:blurnoiseparams}.
For training we use 5000 patches of size $300\times300$ taken from the COCO2017 dataset~\cite{COCO}.
We use the Adam optimizer to minimize the $L_1$-norm between the output and ground-truth images.

Performance evaluation is done using 5000 similarly degraded images from the COCO test split.
Table~\ref{tbl:results} summarizes the quantitative performance of the CNNs -- self-normalization improves CBDNet's ability to restore the images.
Moreover, the lightweight SN-\cbdnethalf\ outperforms the original CBDNet.
Thus, access to a large effective receptive field and accurate estimation of the degradation-parameters is powerful enough to make up for the loss in overall capacity.
Figure~\ref{fig:blurnoiseresults} shows a number of restoration examples.

\begin{table}
	\centering

\newcommand{\spc}{\hphantom{\tiny ,}}%

\small
\begin{tabular}{||c||c|c||c|c||c|c||c|c||}
	\hline
	& \multicolumn{2}{c||}{COCO}						& \multicolumn{2}{c||}{RESIDE}					& \multicolumn{2}{c||}{Cityscapes} \\[-4pt]
	& \multicolumn{2}{c||}{\scriptsize blur + noise}	& \multicolumn{2}{c||}{\scriptsize gray haze}	& \multicolumn{2}{c||}{\scriptsize coloured haze} \\[1pt]
	
							& PSNR				& SSIM				& PSNR				& SSIM				& PSNR				& SSIM \\
	\hline					
	\spc\cbdnethalf\spc		& \spc26.42dB\spc	& \spc0.817\spc		& \spc26.82dB\spc	& \spc0.961\spc		& \spc21.77dB\spc	& \spc0.879\spc \\
	\spc CBDNet\spc			& \spc26.46dB\spc	& \spc0.817\spc		& \spc27.06dB\spc	& \spc0.959\spc		& \spc22.26dB\spc	& \spc0.884\spc \\
	\spc SN-\cbdnethalf\spc	& \spc\bf26.57dB\spc& \spc0.820\spc		& \spc28.16dB\spc	& \spc0.967\spc		& \spc25.30dB\spc	& \spc0.928\spc \\
	\spc SN-CBDNet\spc		& \spc26.55dB\spc	& \spc\bf0.822\spc	& \spc\bf28.64dB\spc& \spc\bf0.970\spc	& \spc\bf26.36dB\spc& \spc\bf0.938\spc \\
	\hline
\end{tabular}

	\caption{Quantitative results on the different tasks.}
	\label{tbl:results}
\end{table}

\begin{figure}
	\centering
	\iflivetikz
		\input{SelfNormedCNN_904_DeblurDenoiseResults}
	\else
		\includegraphics[trim=5 0 0 0,clip]{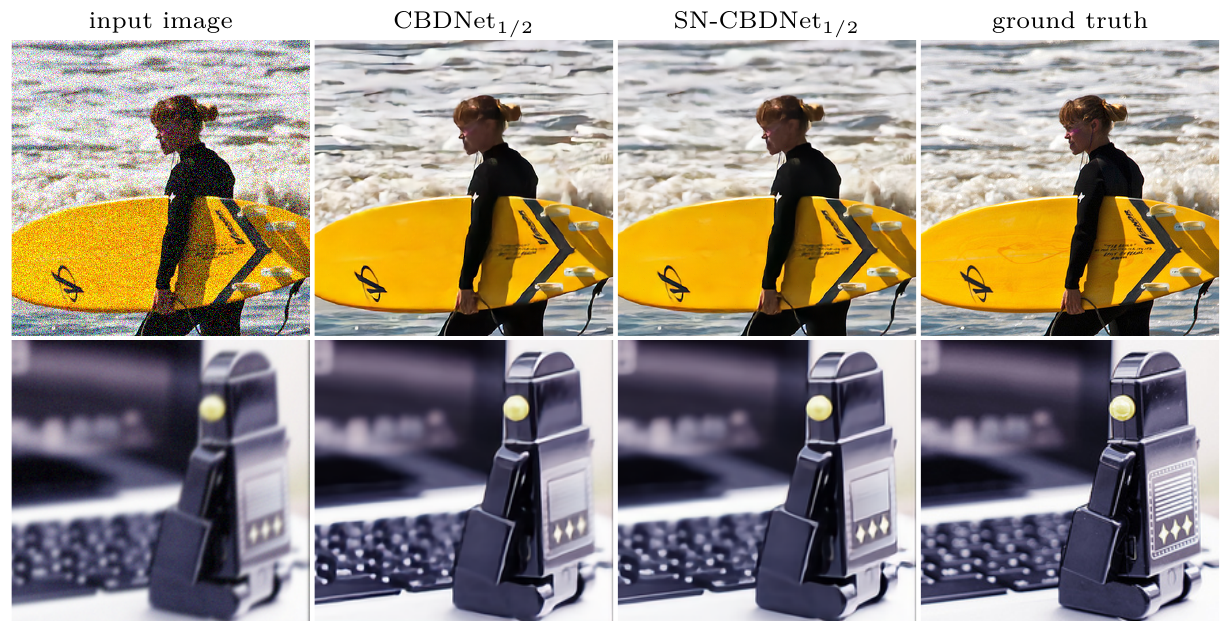}
	\fi
	\caption{Results of the denoise-and-deblur task.
		Top: \cbdnethalf\ performs over-deblurring, creating artefacts in the water, whereas SN-\cbdnethalf\ restores the original details well.
		Bottom: SN-\cbdnethalf\ produces crisper details.}
	\label{fig:blurnoiseresults}
\end{figure}

\subsubsection{Analysis of Self-Normalization}
\label{sec:complex_analysis}

We now inspect in more detail the capabilities of the self-normalization side-chain to estimate degradation-parameters.
All the statistics and examples below are obtained using SN-\cbdnethalf\ on 5000 COCO test images.

Figure~\ref{fig:blurnoise_scatter} shows two-dimensional histograms for each combination of the 5 degradation-parameters versus the 8 state-values of the \sidechain.
One can observe that each of the degradation-parameters is highly correlated with at least one of the coordinates of the state-vector, \ie given the state-vector, a robust estimate of the imaging conditions can be obtained.
Thus, the \sidechain\ provides the main branch of the model with indicators of the current degradation-parameters at an early stage in the processing, and the vast majority of the model's capacity can be invested in the restoration itself.
This is achieved through end-to-end training without any top-down instruction regarding which global features may be beneficial in the given task.

\begin{figure*}
	\centering
	\iflivetikz
		\input{SelfNormedCNN_903_HistComplex}
	\else
		\includegraphics[trim=50 0 0 0,clip,width=1.01\textwidth]{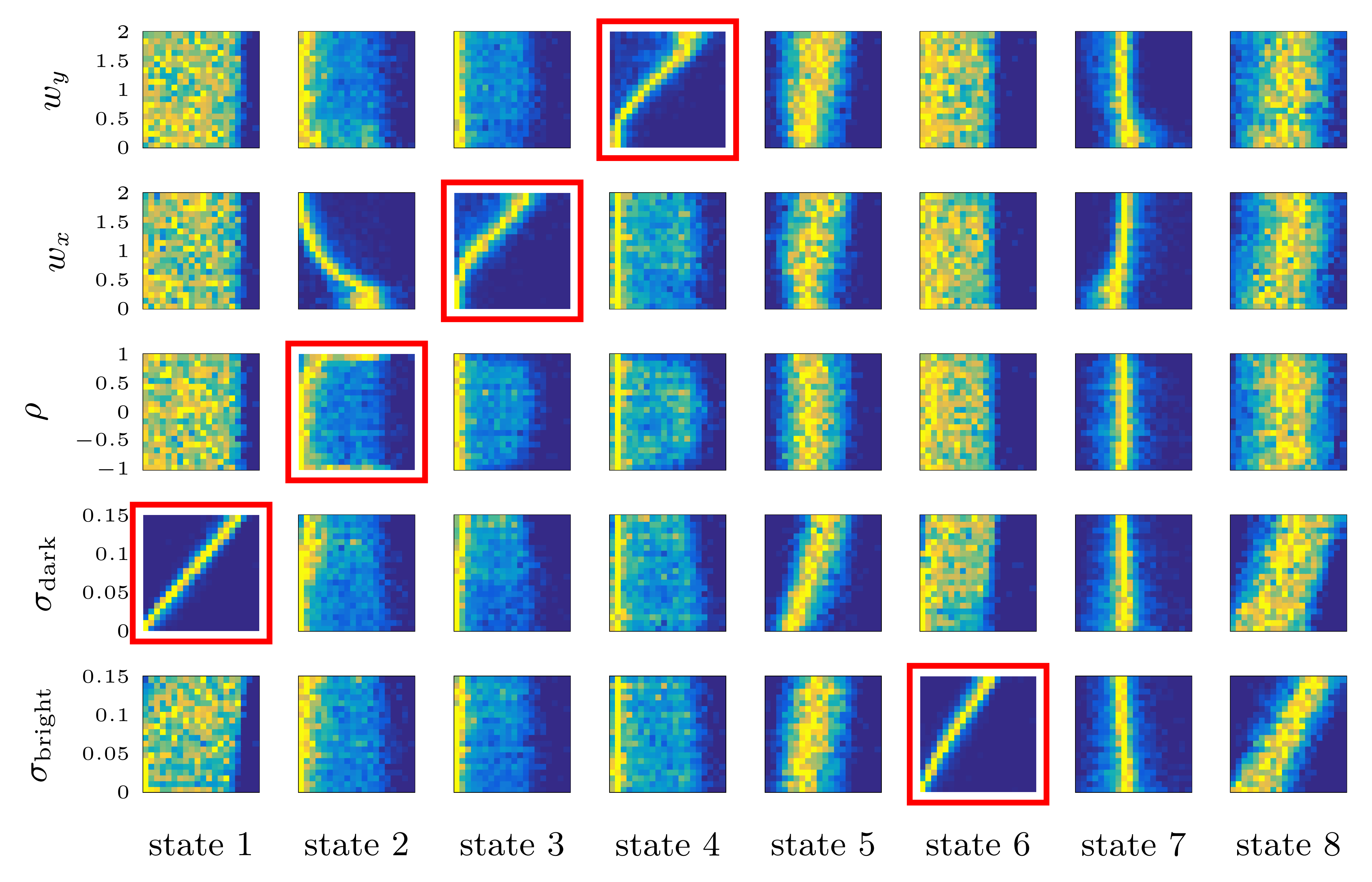}
	\fi
	\caption{Conditional distributions $\text{Pr}\{\,\text{state}\,|\,\text{degradation-parameter}\,\}$ in \mbox{SN-\cbdnethalf} trained for restoration of the multi-dimensional degradation~\eqref{eq:blurnoise} (normalized for clearer visualization).
	For each of the degradation-parameters, at least one state vector can serve as a good estimator (red boxes).}
	\label{fig:blurnoise_scatter}
\end{figure*}

Figure~\ref{fig:validity_relevance} shows a selection of \validity\ and \relevance\ output maps.
The validity maps highlight zero versus non-zero values, whereas the relevance maps are full-range heat-maps.
We observe that the vast majority of pixels have zero validity, \ie they are \emph{not} used in the weighted averaging -- only the most indicative pixels take part in producing the global estimate.
For example, only salient vertical edges are used to estimate the horizontal blur $w_x$.
In the relevance maps we see that the estimates are used in much wider regions of the images.
Thus, an important facet of the functionality of the \sidechain\ is revealed -- information is transported from regions where parameter-estimation is possible to where the parameter is needed.

\begin{figure*}
	\centering
	\iflivetikz
		\input{SelfNormedCNN_905_ValidityRelevance}
	\else
		\includegraphics[trim=5 0 0 0,clip]{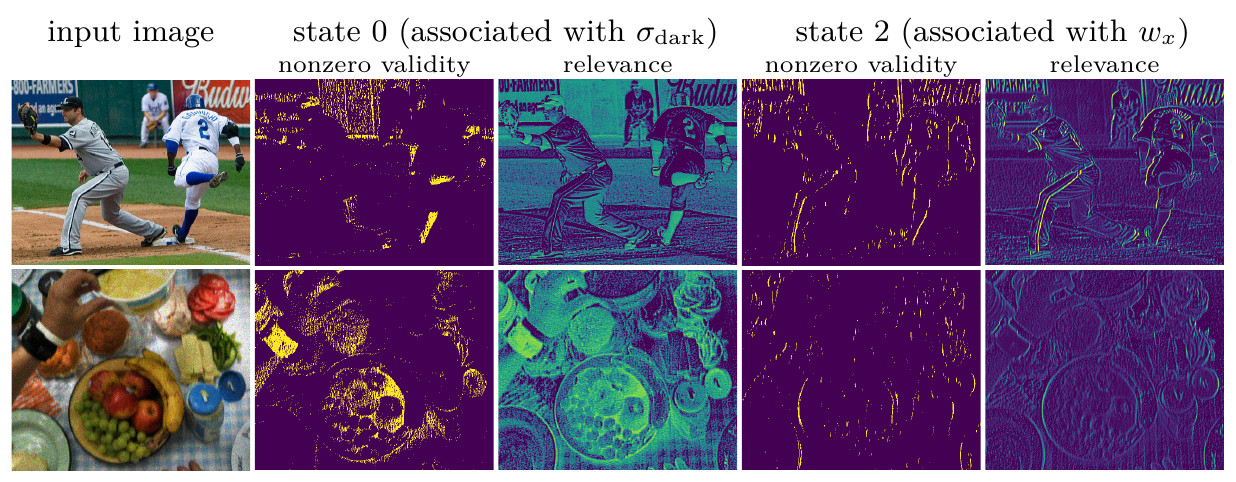}
	\fi
	\caption{Outputs of the \validity\ and \relevance\ components of the \sidechain\ for the state coordinates associated with $\sigma_{\text{dark}}$ and $w_x$.
	The validity maps show zero versus non-zero values.}
	\label{fig:validity_relevance}
\end{figure*}

\subsection{Blind Dehazing}
\label{sec:dehaze}

Removal of haze from images is another image restoration task which is heavily influenced by varying imaging conditions of frame-wide scope.
The prevailing degradation model in the literature is the \emph{atmospheric scattering model}~\cite{DehazeSurvey}, given by
\begin{equation} \label{eq:haze_model1}
I_{\text{hazy}}(x, y) = T(x,y) \cdot I_{\text{groundtruth}} + \left(1 - T(x,y) \right) \cdot A
\end{equation}
where the \emph{transmission} $0 \leq T(x,y) \leq 1$ is a function of the distance of the object in view
\begin{equation} \label{eq:haze_model2}
T(x,y) = \exp \left\{ -\beta D(x,y) \right\},
\end{equation}
and the \emph{air-light vector} $A$ represents the hue of the scattering induced by the imaging medium.

The above model has four global parameters -- the \emph{haze-power} $\beta$ and the three components of the air-light $A$.
An effective dehazing algorithm must be able to estimate, either explicitly or implicitly, these global degradation-parameters, in order to effectively restore the underlying image.
Indeed, previous deep dehazing efforts~\cite{cai2016dehazenet,li2017aodnet,zhang2018dcpdn,Li2019lapnet,Deng_2019_ICCV} all rely on elaborate custom topologies incorporating a global estimation component.
We compare the performance of the four CNNs above in dehazing two datasets, and show that the \sidechain\ has the ability to enable a simple topology, not designed for specifically for dehazing, to perform the task.

RESIDE~\cite{RESIDEbenchmark} is a contemporary dehazing benchmark.
It includes a large corpus of images, each artificially hazed with several values of $\beta$, but restricted to grey air-light $A=(1,1,1)$.
It is not divided into training/validation/test splits -- we selected at random 5000 images for training, 1000 for validation and another 5000 for testing.
For each image we selected one of the $\beta$ values at random.

Cityscapes~\cite{Cordts2016Cityscapes} is a general-purpose dataset, which includes depth information.
We use the depth maps (preprocessed using the technique from~\cite{SDHV18}) to synthesize hazy images with variable $\beta$ and $A$, \ie the degradation-space is four-dimensional.

Example results are shown in Figures~\ref{fig:dehaze_results_cover} (RESIDE) and~\ref{fig:dehaze_results} (Cityscapes).
In Figure~\ref{fig:dehaze_results_cover} we see that CBDNet, having no capability to estimate the frame-wide degradation-parameters, produces non-uniform results, whereas SN-CBDNet avoids these artefacts.
In the case of coloured haze (Figure~\ref{fig:dehaze_results}), the \sidechain\ is able to estimate the air-light hue and reproduce colours with higher accuracy.
\ifappendix
Further analysis is given in Appendix~\ref{sec:dehaze_analysis}.
\else
We provide further analysis of this task (similar to Figure~\ref{fig:blurnoise_scatter}) in the appendices.
\fi
Quantitative results on the test data, given in Table~\ref{tbl:results}, reveal that the value of self-normalization is apparent in dehazing as well.
Again we see that 
the slim SN-\cbdnethalf\ outperforms the hefty CBDNet by a large margin.

\begin{figure*}
	\centering
	\iflivetikz
		\input{SelfNormedCNN_909b_CityscapesResults}
	\else
		\includegraphics[trim=5 0 0 0,clip]{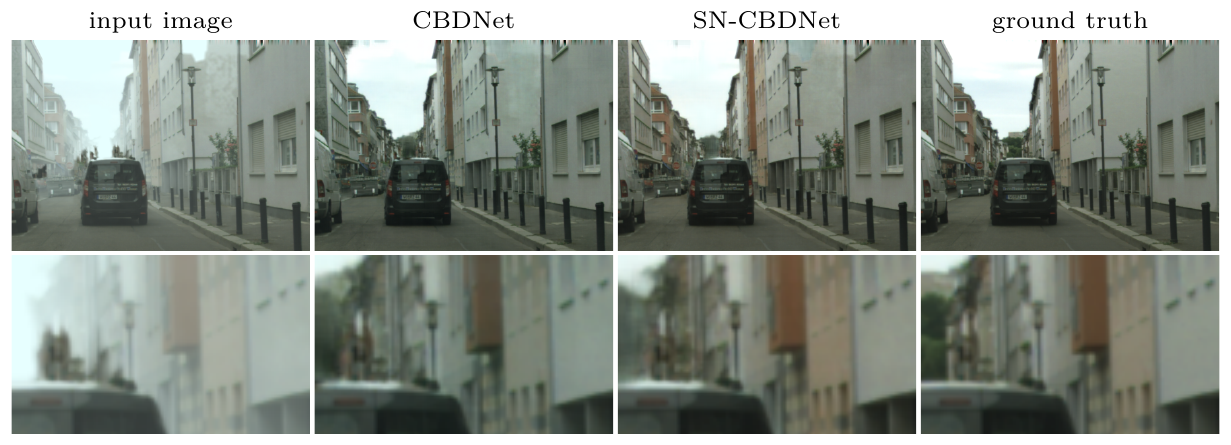}
	\fi
	\caption{Dehazing example from the Cityscapes dataset.
		Under blueish haze \mbox{SN-CBDNet} exhibits superior colour reproduction.}
	\label{fig:dehaze_results}
\end{figure*}

\subsection{Using the \sidechain\ for Objective-Tuning}
\label{sec:flexible}

Blind restoration models which employ an explicit parameter-estimation mechanism (\eg~\cite{CBDNet,GRDN}) possess the potential for influencing the output of the CNN by transforming the estimated values before their utilization in the main processing model.
Self-normalized CNNs also have this potential, which can be realized by manipulations on the \relevance\ component of the \sidechain.

We created a variant of SN-\cbdnethalf\ that has two instances of the \relevance\ component of the side-chain, which can be toggled using an external control (Figure~\ref{fig:side_chain_switch}).
This variant was trained for the task in Section~\ref{sec:complex}, but instead of using $I_{\text{groundtruth}}$ as the target image, we toggled between $I_{\text{groundtruth}}$ and $I_{\text{blurred}}$, while also toggling between the two \relevance\ instances.
In order to produce an image close to $I_{\text{blurred}}$, the model must perform only denoising aspect of the task, without the deblurring, \ie the trained model has \emph{two distinct modes of operation}.
Note that the CNN is trained in parallel for both modes -- during the training, each mini-batch includes half of each kind.

\begin{figure}
	\centering
	\iflivetikz
	\input{SelfNormedCNN_902b_SideChainInternal_double}
	\else
	\includegraphics[trim=0 0 0 0,clip,width=\textwidth]{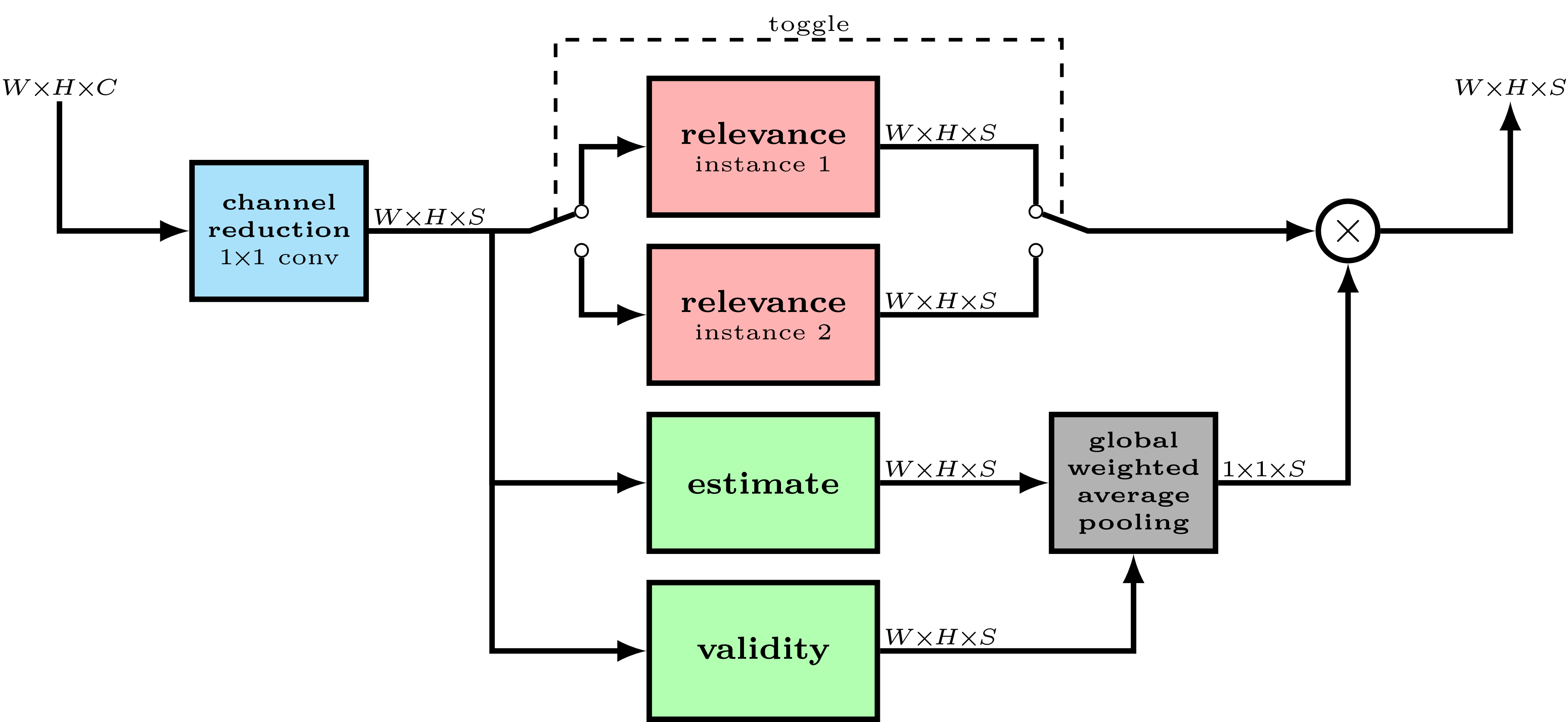}
	\fi
	\caption{Variant of the Self-Normalization Side-Chain for toggling between operation modes of the restoration CNN.}
	\label{fig:side_chain_switch}
\end{figure}

At inference, this new model can be switched between the two modes of operation -- simultaneous denoise and deblur (as before) or denoise only -- by selecting which of the \relevance\ instances is used.
All the trained parameters outside the two \relevance\ components are joint.
In particular, the other components of the side-chain are used in both modes of operation -- \ie the same imaging conditions are estimated, but applied differently.
We provide example results in Figure~\ref{fig:switch}.

\begin{figure*}
	\centering
	\iflivetikz
		\input{SelfNormedCNN_908_Switchable}
	\else
		\includegraphics[trim=5 0 0 0,clip]{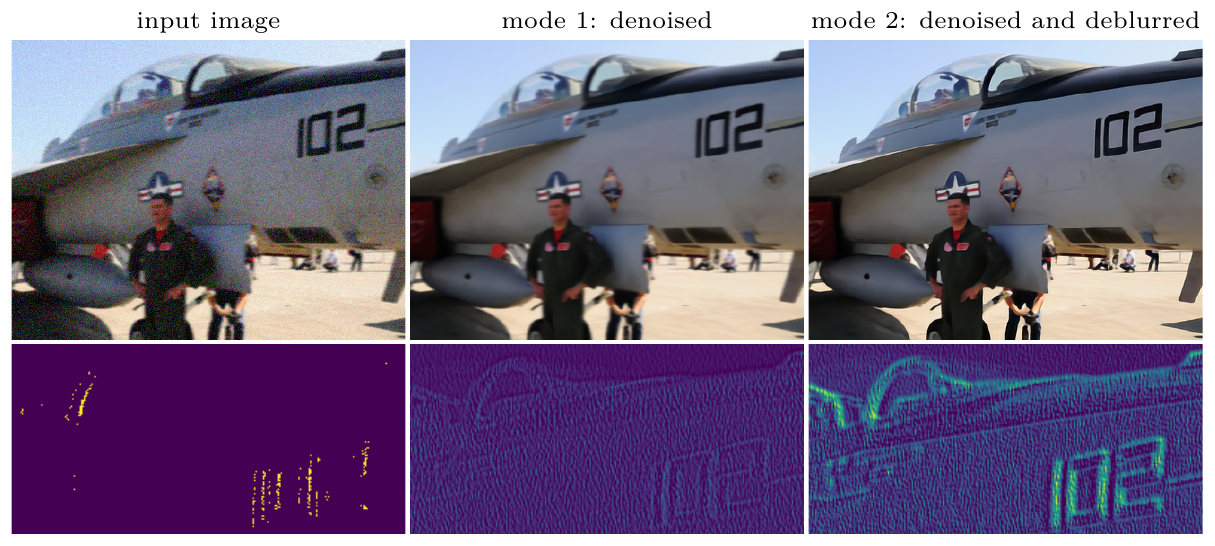}
	\fi
	\caption{Demonstration of the switchable variant of SN-\cbdnethalf.
	Bottom: \validity\ (left) and the two \relevance\ maps associated with horizontal blur $w_x$.
	In mode 1 (denoise only) the \relevance\ is very low, effectively ``blocking'' the blur-measurement from reaching the main CNN.}
	\label{fig:switch}
\end{figure*}

Contrary to most tunable-CNNs (\eg~\cite{DeepInterp,Shoshan_2019_ICCV}), the overhead -- an extra instance of the \relevance\ -- is negligible.
We plan to dive deeper into this potential in future work.

\section{Conclusions}
\label{sec:conclusions}

We have introduced a trainable architectural component designed for estimating global imaging conditions, while using only end-to-end training with no prior knowledge of the task.
This component can be added to most existing CNN topologies.
We have demonstrated and analysed its capabilities in several different image restoration use-cases.

Avenues for future research include:
\begin{itemize}
\item Testing utility of the \sidechain\ in scene-understanding tasks under harsh conditions (noise, night-time, haze, etc.).
\item Augmenting the \sidechain\ to take into account the non-uniform influences on global imaging conditions (\eg the effect of vignetting on noise-distribution across the frame).
\item Determining the required capacity $S$ of the state-vector.
\item Advancing the technique explored in Section~\ref{sec:flexible} to produce continuous parametrization of the CNN output.
\end{itemize}

\fi 

\ifappendix

\ifbody
\clearpage
\fi
	
\appendix

\section{Analysis of Side-Chains in Dehazing Models}
\label{sec:dehaze_analysis}

\begin{figure}
	\centering
	\iflivetikz
		\input{SelfNormedCNN_910_HistDehaze}
	\else
		\includegraphics[trim=50 0 0 0,clip,width=\textwidth]{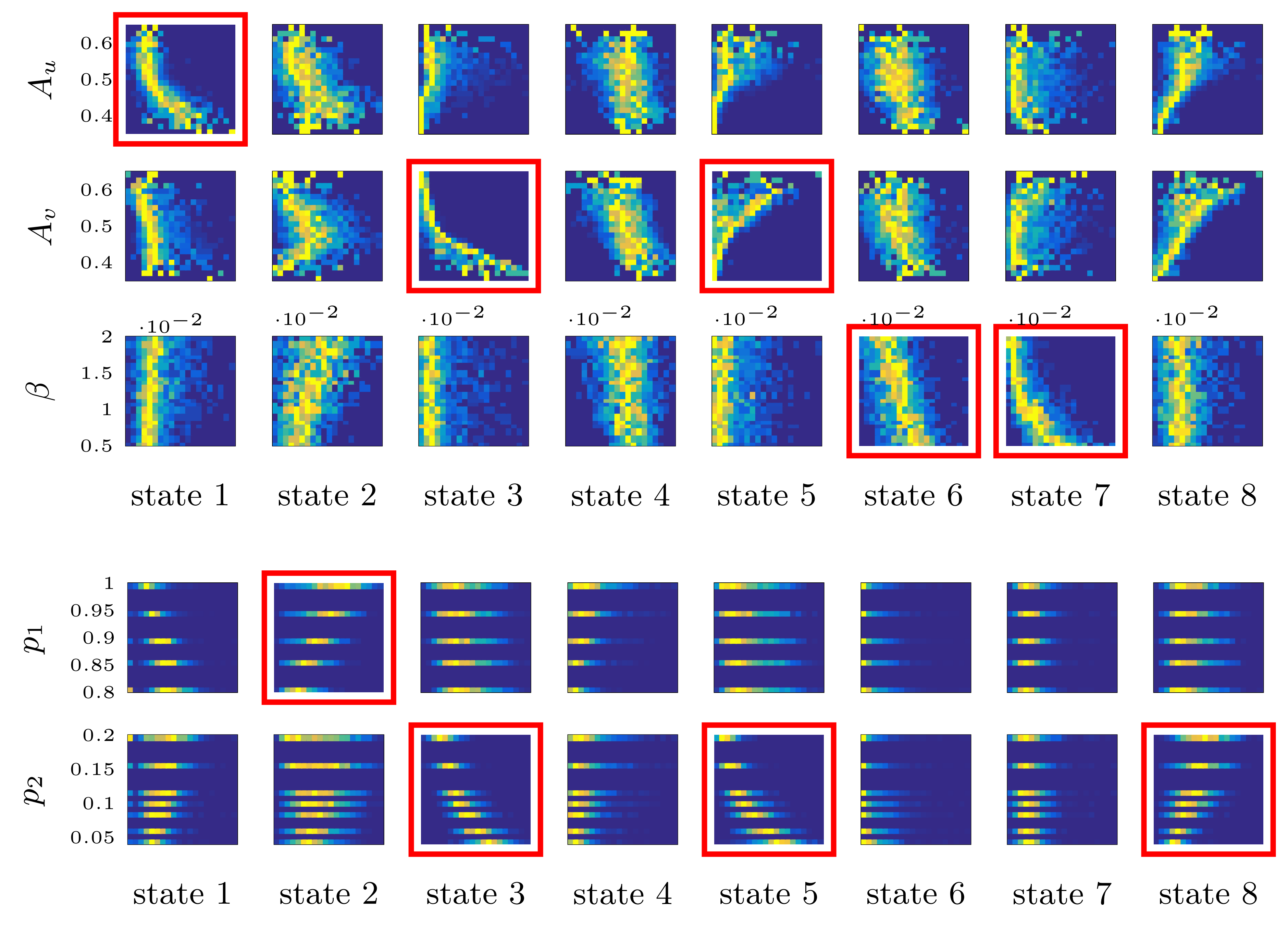}
	\fi
	\caption{Conditional distributions $\text{Pr}\{\,\text{state}\,|\,\text{degradation-parameter}\,\}$ in \mbox{SN-CBDNet} trained on the two dehazing datasets.
	Top: Cityscapes with hued air-light. The parameters are the U and V channels of the air-light vector $A$ in YUV representation, and the haze power $\beta$ in Equation~\eqref{eq:haze_model2}.
	Bottom: The RESIDE benchmark. The file-names of the input images have two numbers describing the synthesized haze -- we could not find documentation of the meaning of these two values, but the \sidechain\ was nonetheless able to deduce their importance to the task from the data itself.}
	\label{fig:dehaze_scatters}
\end{figure}

\section{Block Diagram of SN-CBDNet}
\label{sec:sn_cbdnet}

Figure~\ref{fig:sn_cbdnet} is a diagram of the self-normalized CBDNet.
Layer naming is based on the official \texttt{MatConvNet} implementation published by the authors of~\cite{CBDNet}.
The components which differentiate \mbox{SN-CBDNet} from the baseline CBDNet are marked in red -- the \sidechain\ itself, channel concatenation, and a point-wise convolution to fuse the output of the \sidechain\ and the main signals.

The topologies of \cbdnethalf\ and SN-\cbdnethalf\ are identical, but they have half the number of channels in all layers.

\begin{sidewaysfigure}
	\centering
	\iflivetikz
		\input{SelfNormedCNN_911_SelfNormedCBDNet}
	\else
		\includegraphics[trim=0 0 0 0,clip,scale=0.875]{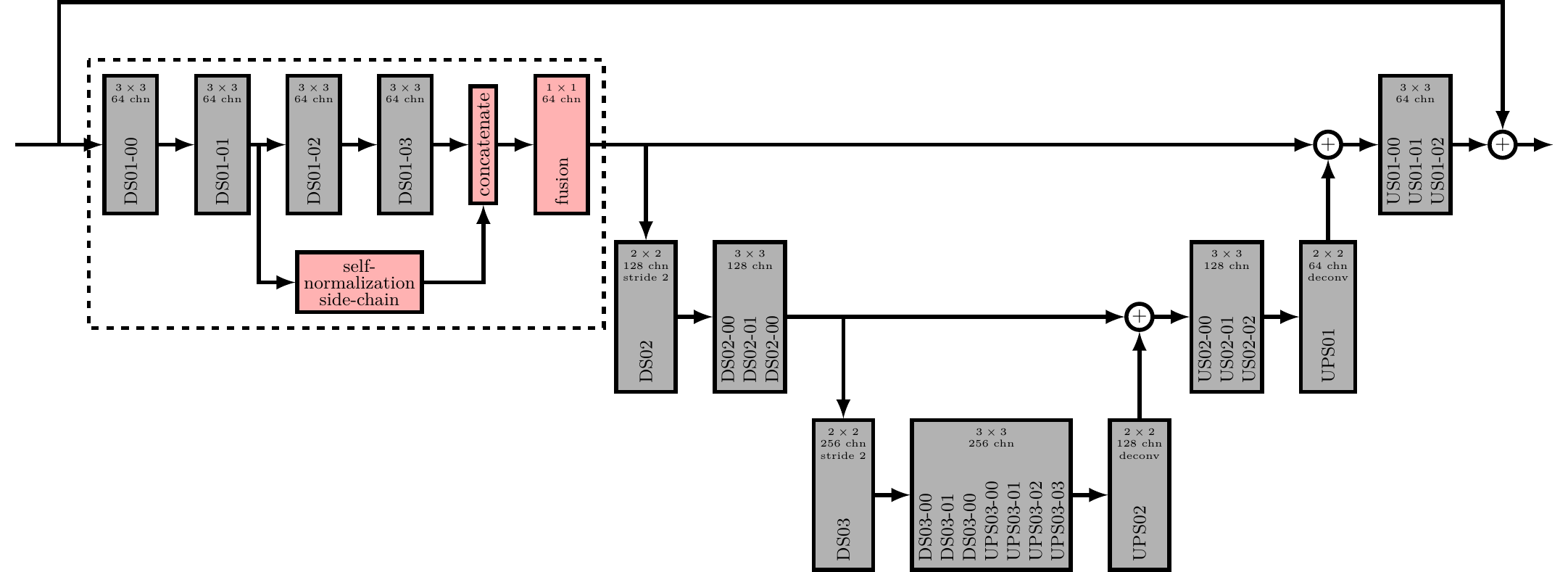}
	\fi
	\caption{Self-normalized CBDNet.
	The \sidechain\ (red) is attached to the full-resolution encoder (dashed rectangle), which is outlined in detail.
	The rest of the UNet is drawn more compactly -- simple sequences of identical layers are grouped.}
	\label{fig:sn_cbdnet}
\end{sidewaysfigure}

\fi 

\clearpage
%
%
\bibliographystyle{splncs04}
\bibliography{SelfNormedCNN}
\end{document}